**Discursive objection strategies in online comments: Developing a classification schema and validating its training**

Ashley Shea, 165 Mann Library, Cornell University, Email: ald52@cornell.edu
Aspen K.B. Omapang, Cornell University
Ji Yong Cho, Cornell University
Miryam Y. Ginsparg, Cornell University
Natalie Bazarova, Cornell University
Winice Hui, Cornell University
René F. Kizilcec, Cornell University
Chau Tong, University of Missouri
Drew Margolin, Cornell University

Abstract

Most Americans agree that misinformation, hate speech and harassment are harmful and inadequately curbed on social media through current moderation practices. In this paper, we aim to understand the discursive strategies employed by people in response to harmful objectionable speech in news comments. We conducted a content analysis of more than 6,500 comment replies to trending news videos on YouTube and Twitter and identified seven distinct discursive objection strategies (Study 1). We examined the frequency of each strategy's occurrence from the 6,500 comment replies, as well as from a second sample of 2,004 replies (Study 2). Together, these studies show that people deploy a diversity of discursive strategies when objecting to speech, and reputational attacks are the most common. The resulting classification scheme accounts for different theoretical approaches for expressing objections and offers a comprehensive perspective on grassroots efforts aimed at stopping offensive or problematic speech on social media.

This research was funded by the National Science Foundation Division of Information and Intelligent Systems (NSF, funding number: 2106476). It was accepted and presented at the International Communication Association's Annual Meeting in Toronto, ON May 2023

Introduction

Platforms like Youtube, Twitter, Reddit and Facebook are ubiquitous parts of daily information consumption, affording low-barrier opportunities to engage with people and content. User-generated comments that appear below uploaded content on social media platforms play an important role in shaping our understanding of content, and serve as important credibility cues that users incorporate into their judgments about the original content (Kwek et al., 2020; Sundar, 2008). While online comments sections have been recognized for their role in enabling social connectivity (Wu & Atkin, 2017) and promoting deeper engagement with news stories and the journalists writing them (Meyer & Carey, 2014), comment sections have also been criticized for providing opportunities to generate and perpetuate racial hostility (Hughey & Daniels, 2013), political violence (Wahlström et al., 2021) and disinformation (Petit et al., 2021). In the two studies we introduce below, we explore the authentic real-world discursive strategies that users employ to deter an objectionable comment they have encountered online and the frequency of their occurrence. *Objectionable content* refers to the harmful and problematic content that users want to deter in online comment threads. This includes content that people consider abhorrent or dangerous, such as hate speech, conspiracies, and disinformation, but also things like ideologically opposing viewpoints and comments about detested politicians and their policies.

While nearly 70% of Americans state that social media has had a negative influence on civility in the U.S. (Auxier, 2020), use of social media platforms and their features remain very popular. Of the 80% of U.S. adults that report using YouTube, nearly 70% report daily use of the platform (Auxier et al., 2021). Deterring objectionable content on such popular social media platforms is a crucial goal for cultivating a healthy and deliberative digital public sphere.

Social media platforms and websites with interactive features have introduced a variety of measures to limit the spread of objectionable content within comment threads. Some have completely dismantled comment sections (Labarre, 2014), and others have tried to promote their promise for long-form idea-exchange by limiting participation to credentialed users (Reardon, 2013). Platforms have introduced individual tools like flagging and reporting, integrated algorithmic solutions for flagging and classifying content, and have recruited content moderators to assist with verification and removal processes.

Due to the variance among individual perceptions of what constitutes objectionable content, platform-driven solutions are inadequate. Users inevitably encounter content they consider objectionable, with some users then forming their own response to such content.  How everyday users try to counteract and oppose objectionable content is important to understand. The identification of different types of discursive objection strategies and their frequency of use on different platforms contributes to understanding of the participatory process through which social media communities attempt to regulate content. This research opens the door for future studies on the efficacy of different types of discursive objection strategies on audiences and offenders, as well as training opportunities to empower users to object effectively to promote civil discourse on social media.

To date, little research exists on user counter strategies in comment threads. In recent years, several articles have explored how social media users have tried to dole out punishment in the name of so-called justice when playing witness to racist comments (Blackwell et al., 2018; Ronson, 2015) but more work is needed. The goal of our research is to address this gap and develop a comprehensive coding scheme for classifying types of discursive objection strategies used in online comments. For the purpose of this work, we introduce the concept of a discursive objection strategy as a persuasive or punishing tactic used to confront and deter objectionable content in online comments.

Literature review

Considerable work has examined when people choose to speak up (Yu et al., 2021; Zerback & Fawzi, 2017; Zhu & Fu, 2021) under what conditions (Xiao & Su, 2022) and in response to what content (Freelon et al., 2016). The two studies that we introduce below complement prior work by focusing on how people try to deter and suppress future objectionable comments on social media once they have decided to speak up.

There are three key approaches to deterring offensive content in social media: 1.) prosecution of online hate crimes, 2.) technological solutions for content flagging and removal, and 3.) mobilizing individuals to engage in counter-speech (Blackwell et al., 2018; Blaya, 2019). Prosecution of online hate crimes is rare. While exploration into the legal loopholes that prevent hate speech from being prosecuted (Levin, 2002; Perry & Olsson, 2009) and the creation of frameworks to aid in the development of new legislation (Bakalis, 2018) are helpful, they are insufficient. Adoption of such frameworks takes time, requiring multilateral coordination across international justice systems. Algorithmic regulation is effective in conjunction with human verification (Gillespie, 2018), but this deterrence strategy is also imperfect. When top-down technocratic solutions try to reduce complex social phenomena into structured equations (Janssen & Kuk, 2016), false positives are inevitably flagged for removal while true positives are missed and remain. Individual users offer an important layer of defense, and while some media and information literacy training programs have focused on individual responsibility (Jeong et al., 2012; Koltay, 2011; van der Meer & Hameleers, 2021), they remain under-utilized. As humans, users have the ability to detect nuance and distinguish between harmful and innocuous content in ways that algorithms cannot. Such interpretations are honed by internalized senses of social and cultural norms (Gavrilets & Richerson, 2017).

There are a number of different ways that individuals can intervene when they encounter objectionable content within an interactive comment thread. Broadly, user-level discursive strategies can be characterized into two types: 1) comments that *punish* and 2) comments that try to *persuade*. User comments that punish can be further distinguished into three primary strategies: face threats (Brummernhenrich & Jucks, 2016; Oeldorf-Hirsch et al., 2017), public shaming (Basak et al., 2016; De Vries, 2015; Hou et al., 2017; Klonick, 2015; Koganzon, 2015; Surani & Mangrulkar, 2021) and online disassociation (Kádár et al., 2013; Synnott et al., 2017). Face threats refer to criticism of a person's behavior or attitudes and are generally meant to upset the recipient (Cupach & Carson, 2002). Public shaming is similar, but condemns someone for their violation of social norms in a public way to arouse embarrassment and remorse (Basak et al., 2019). Online disassociation is a strategy used to ostracize the offender, often used to point out the offender's flawed behavior. Generally, all try to punish the "wrongness" of objectionable content by creating social distance between the offender and offended. No single review has brought these three strategies together to synthesize the commonalities and distinctions of punishing strategies.

Another way to object is to try to persuade the community, rather than threaten with sanctions (Guadagno & Cialdini, 2009). Broadly there appear to be two primary persuasion strategies that attempt to deter objectionable content by encouraging alternative, more desirable behaviors on social platforms. The strategies include use of moral reasoning (Jenkins et al., 2012), and use of counter-narratives (Bartlett & Krasodomski-Jones, 2015; Braddock & Horgan, 2016). Moral reasoning plays on emotion and logic by arguing to do what is in the best interest of the greatest number of people (Hunter, 1974). Social norm nudging is a form of moral persuasion and exposes users to information about "typical" attitudes or behaviors that might differ from their own, framing these attitudes and behaviors as desirable and worthy of believing and doing (Masur et al., 2021; Rost et al., 2016; Spottswood & Hancock, 2017).

Counter-narratives are messages that push back on the dominant or visible story to encourage a different way of thinking (Herman, 2007). Counter-narratives can utilize emotion and logic to challenge a particular paradigm, but can also use amplification techniques like sound boosting and visibility markers to drown out objectionable content. Hashtags are common counter strategies used on social media that rely on amplification to persuade. When  #AllLivesMatter and #BlueLivesMatter emerged as counter hashtags to #BlackLivesMatter, their use was meant to overshadow one view (that Black Lives Matter) by amplifying opposing views (that Blue Lives and All Lives Matter) (Chen & Berger, 2013; Gallagher et al., 2018; Masullo Chen et al., 2018). Similar to punishment strategies, no single review has synthesized the commonalities and distinctions between persuasion strategies used to deter objectionable content on social platforms.

Many studies have explored the ways in which harmful content materializes and can be classified and programmatically detected for removal. Various typologies have been developed to understand and detect the granularity of hate speech (Alorainy et al., 2019; Cinelli et al., 2021; Wahlström et al., 2021), disinformation (Sharma et al., 2022; Zhang & Ghorbani, 2020), and toxicity and harassment in online discussion forums (Andročec, 2020; Blackwell et al., 2017; Rahul et al., 2020). The present study is the first to operationalize the broader construct of *objectionable content* to explore all relevant forms of problematic content encountered in user-generated comments on two social media platforms. This research is also the first of its kind to collate and categorize together the distinct discursive strategies that people are actually using on social media when objecting to such objectionable content in comment threads. Our classification schema identifies and defines each strategy type, and includes the unique defining features of each.

Our approach consists of two studies: the first to develop and validate a comprehensive classification of discursive objection strategies derived from comments on social media platforms (Twitter and YouTube) and the second is to examine the relative frequencies of the validated objection types sampled from comments appended to videos on YouTube.

STUDY 1: Developing a codebook

For our first study, we ask the following research questions:

RQ1 What are the discursive strategies that users employ when objecting to online comments and do these strategies fit within the broader strategies of punishment and persuasion from the relevant literature?

RQ2 What are the defining features of each discursive strategy used to object?

METHODS

Codebook Development

To develop a classification schema of discursive objection strategies in our first study, our team followed a two-phase process. In the first phase, we followed the rigorous 6-step approach of collaborative thematic analysis (Richards & Hemphill, 2018) in which we embarked on 1) preliminary planning and sampling, 2) open & axial coding,  3) preliminary codebook development; 4) holistic sorting; 5) internal testing and 6) finalization of a codebook. At the conclusion of the first phase, we had a 7-strategy classification schema ready to validate in the second phase.

In the second phase, we validated the codebook using crowdworkers from Amazon's Mechanical Turk ("MTurkers") who were trained in applying the coding scheme to real comments. MTurkers were fed a collection of curated comments previously sampled and coded in the first phase and the accuracy of their responses was measured against the "ground truth" established by our team. Phase 1 and phase 2 processes of the first study are illustrated in Figure 1 below.

Figure 1. Study 1: Codebook Development & Validation Process

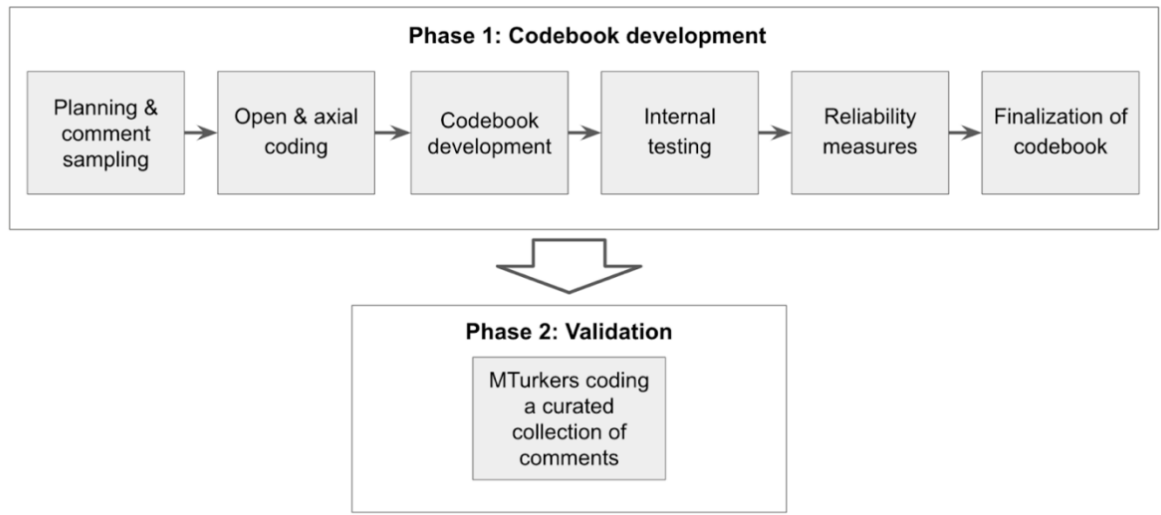

Comment Sampling

Our sample of comments for Phases 1 and 2 was derived from two prominent social media platforms, YouTube and Twitter. There is little empirical work examining comments on YouTube, in part due to technical limitations in accessing the nested data, but its popularity across race, age and gender (Auxier & Erson, 2021) makes it an important platform to study. We selected Twitter as a second platform to include in our sampling due to its accessible API, and the threaded discussion it enables. Across both platforms, we limited our sample to comment threads with multiple replies in hard news topics (e.g., war, medicine, science, etc.). Discussion of hard news is generally viewed as more divisive (Lehman-Wilzig & Seletzky, 2010; Schramm, 1949) and therefore we anticipated that videos on hard news topics would have a larger number of argumentative comment threads, filled with the very objection strategies that we wanted to examine.

To collect our preliminary sample from YouTube, we utilized the Google Trends API to identify the topics that users searched most frequently on the platform. After limiting the search topics to hard news only, we were left with ten topics to include in our analysis. To move from our topics to a corresponding video for each, we utilized YouTube's "News" tab within the YouTube-curated "Explore" section where a selection of highly viewed content from popular content creators are algorithmically curated. We selected the video per topic with the most views, and then scraped the top 100 comments and their reply chains for each video.

Twitter does not have a comparable mechanism to identify topics that users are most interested in. Trending topics are fleeting and it is often easier to locate high-engagement conversations by focusing on controversy instead of news topics. To identify timely controversial subjects, we utilized Reddit's popular r/CMV (change my view) subreddit where users are prompted to discuss current controversial topics based on moderator and user suggestion. Topics from Reddit were then searched on Twitter to identify Twitter threads containing at least three comments.

Open & Axial Coding and Codebook Development

After acquiring a comment sample of 7,500 comments for Phase 1, we embarked on the next step of our collaborative thematic analysis: open and axial coding. The purpose of open and axial coding was to: 1) identify categories that objection strategies could be sorted by, 2) distinguish between categories, 3) define each category clearly and 4) be sure categories were comprehensive, catching every strategy we encountered. When looking for objections within the comments, we were focused on direct replies to other users, and not replies directed at the video, or the video's sourcing news agency (i.e., CNN). We focused on direct replies to other users because we are interested in the discursive strategies that users employ to persuade or punish other users. A team of four graduate students reviewed user comments in batches of 800 and looked for persuasive or punishing tactics as per the literature reviewed in the previous section. All direct replies that fit into these broad categories were flagged.

Over the span of three months, the team of coders met weekly to discuss potential categories and labels that emerged from independent sorting and labeling. During this phase of sorting and labeling, it was important to draw a distinction between comments with disagreements and comments with objection strategies. Although comments that disagree do include differing opinions and might utilize persuasive or punitive elements, objection strategies do more than disagree. Objection strategies also want to limit the spread of information, silence the user who made the comment and deter similar comments from being made in the future. Deliberative dialogue and negotiation of knowledge - two processes fundamental to healthy civic engagement and community vitality- often requires disagreement. Objection strategies differ from disagreement because people consider the content to which they are objecting so abhorrent or dangerous to community vitality that either its creator's opinion must be changed, their credibility must be undermined, or they must be removed.

Disagreements are common; objection strategies are rare but potentially more impactful on the discussion thread because they attempt to uphold or change prescriptive and/or descriptive norms within the comment space. After collectively coding a total of roughly 7,500 comments from Youtube and Twitter, 575 total objections were identified by our team, representing less than 10% of our comment sample. Eight preliminary objection strategies emerged to sort all 575 objections following this initial coding and the eight objection strategies were ready for further testing.

With eight strategies in a preliminary codebook, the coders received a new batch of 400 comments each and again reviewed only comments that were direct replies to users. To test whether categories were comprehensive, coders sorted comments into one of the eight categories, or into an "other" category if an existing category was inadequate. In this iteration of sorting, less than 5 comments were sorted into the "other" category, and one category did not receive any comments at all. Some categories included many comments, while others had few. Some strategy names were narrowed to become more focused and understandable, while other strategy types were broadened to become more inclusive of comments assigned the "other" label. Two strategy types ("Dismiss objectionable content" and "Content attack") were combined after their similarities proved too confusing, with one not being not being used at all due to its unclear distinction. A newly refined codebook consisting of 7 strategies emerged, and all comments fit neatly into at least one category. In some cases, comments fit into multiple categories if two or more distinct discursive strategies were employed within the same comment. Some categories were rare while others contained many examples. We anticipated unequal distribution of use, and we explore this question of frequency in our second study. For strategy names, definitions and an exemplar comment for each, see Table 1.

Table 1. Classification of objection strategies

| Objection strategy | Description | Exemplar |
|---|---|---|
| Conscience + Imploring | Comments that encourage or chide a person to be better by appealing to that person's conscience. | @User If you are a veteran then you know you can't say that, be proud and a soldier |
| Logic + Imploring | Comments that encourage or chide a person to think differently by using logic or a reasoned argument. | @User better do a little research and go back to the original article because that's not what it said. Its not even logical |
| Intimidate + Violence | Comments that try to intimidate the user that left the comment by directly threatening or implying violence toward them, their "in group" or aligned "leader". | @User Anybody who attempts to ban/outlaw AR-15's or AR-10's from We the People deserves death by the most painful means possible. |
| Smear + Undermine | Comments that attack or "smear" a user's character to undermine their authority. | @User Quacks like a RACIST republican, you get called a RACIST republican |
| Accusatory Label + Reject | Comments that use accusatory labels to attack the content of a comment. | @User FAKE NEWS!!!!! |
| Self-Preserve + Exit | Comments that directly state or signal exit from the conversation, comment thread or platform to preserve time and energy. | @User I'm not talking to you anymore because it's.. just exhausting.. literal EVERYTHING you said is a lie. And you're biased and delusional... We're done. |
| Space-Preserve + Removal | Comments that direct a user to remove themselves or their comment from the conversation to preserve the norms or decorum of a space, comment thread, or platform. | @User Go be ignorant and hateful elsewhere. |

Final internal testing

With the 7-strategy codebook ready for final internal testing, two coders engaged in consensus coding to determine if categories were being applied consistently between two coders. This involved independent coding of a new dataset of 400 comments filtered by direct replies, with each coder individually sorting comments into one or more of the seven categories before comparing category assignments and calculating intercoder reliability.

Krippendorf (Krippendorff, 2004) suggests >= .800 is high reliability and >= .667 is the lowest threshold for reliability, though these thresholds were published well before the popularization of large-scale social media research, where the use of Krippendorf's Alpha on social media data has had mixed results. Whereas a study of political tweet sentiment using a binary variable achieved an alpha of 0.79 (Monti et al., 2013), the assessment of Twitter profiles as bots also using a binary variable yielded an alpha of 0.51(Im et al., 2020), illustrating the variability in alphas for large-scale social media research.

The results of our testing were 88% agreement and a Krippendorf's Alpha of >=0.50. Given the final number of categories (7) and rarity of objections in samples (~10% on average), our coefficient of 0.88 indicated fairly strong agreement between coders, enough to complete the first phase of our project (Zhao et al., 2013). We finalized the 7-category schema of objection strategies and moved onto the second phase of our study to verify its validity using crowdworkers on Amazon's Mechanical Turk who were trained in applying the coding scheme to real comments from Youtube and Twitter.

Validating the Codebook

Participants

Adult survey respondents 18 years or older were recruited from the crowdsourcing website Amazon Mechanical Turk using the MTurk Toolkit on CloudResearch. CloudResearch is an online study management tool that provides infrastructure to create research tasks that can be crowdsourced while enabling links to externally-hosted surveys like Qualtrics. Respondents were paid $2.20 for the tutorial and quiz plus $2.20 additional if they completed all questions on the test.

Past studies on Amazon Mechanical Turk have raised concerns about the quality of data collected from online workers geographically dispersed (Paolacci et al., 2010), but recent studies have demonstrated that screening out participants with low approval ratings serves as a sufficient mechanism for ensuring high quality data (Peer et al., 2014). Approval ratings are the composite and adaptive scores derived from each user's task completion rate on the platform and their percentage of tasks that get "approved" by each task's creator. When recruiting study participants, we restricted participation to MTurk workers with a high approval rating (95% or above approval ratings). To ensure English language competency, we also restricted our recruitment to MTurk workers that are living in the U.S.

Procedure

Individual participants were trained and tested on only one of the seven objection strategies. This study design was informed by research on the counter-framing effects of user comments (Liu & Mcleod, 2019) in which different counter-framing approaches (alternative framing or direct challenging) were treated as separate conditions paired with distinct tones of response (no verbal attack, verbal attack, or uncivil verbal attack). In our study, participants were randomly assigned to one of the seven trainings, with each training consisting of informed consent, a tutorial to define an objection strategy and illustrate its use in practice (see appendix), a 6-item quality-check quiz, and for those that correctly answered at least 5 of 6 questions on the quiz, a test of 8 additional questions pre-coded by our research team. In the test, MTurkers were provided with random comments from the pre-coded sample and answered either yes or no to whether each comment utilized the learned objection strategy.

RESULTS

A total of 371 MTurkers were recruited and 234 (63.07%) proceeded to the test. The mean quiz pass rate ranged from 34.55% to 82.35%, indicative of the range of difficulty that accompanies strategies that vary in their visibility and distinction, but also indicative of the range of quality among initial survey respondents. Of the 234 that took the test, the average mean test scores for respondents that passed the quiz ranged from 83% to 94%.

Among test takers within each condition, the average agreement ranged from 89% to 98% with an average of 3 coders assigned to each comment. The findings indicate that the training works very well in teaching users to understand a specific objection strategy and identify its use in practice.

**Table 2.** *Results of Mturkers Coding a Curated Collection of Comments*

| Category | Recruitment | Final sample for test | Quiz pass | Avg. quiz score | Avg. coders[a] | Avg. agreement | Reliability[b] |
|---|---|---|---|---|---|---|---|
| Label + Reject | 112 | 49 | 43.75% | 4.15 | 3.92 | 0.92 | 0.37 |
| Imploring + Logic | 110 | 47 | 42.73% | 4.3 | 3.76 | 0.98 | 0.37 |
| Imploring + Conscience | 58 | 27 | 46.55% | 4.33 | 2.16 | 0.89 | 0.26 |
| Self-preserving + Exit | 49 | 38 | 77.55% | 4.98 | 2.88 | 0.92 | 0.04 |
| Space-preserve + Removal | 49 | 34 | 69.39% | 4.88 | 2.72 | 0.89 | 0.12 |
| Smear + Undermine | 40 | 25 | 62.50% | 5.42 | 2.96 | 0.89 | 0.44 |
| Intimidate + Violence | 45 | 39 | 86.67% | 5.22 | 3.07 | 0.96 | NA |

*Note*. [a]The average number of coders per comment, [b]Krippendorf's alpha was calculated. High agreement results in NA.

STUDY 2

In our second study, we asked the following question:

RQ3 What are the relative frequencies of objection types in comments?

For study 2 we report the results from three phases of work. First, we report the relative frequencies of objections in our original dataset of 575 objections. Specifically, out of these 575 we identified 11 of Intimidate + Violence (1.9%), 18 of Space-preserve + Removal (3.1%), 30 of Self-preserving + Exit (5.2%), 54 of Imploring + Conscience (9.4%), 82 of Label + Reject (14.2%), 147 of Imploring + Logic (25.6%) and 223 of Smear + Undermine (38.8%).

While these statistics provide a rough sense of the relative frequency of different objection types (e.g. that intimidating with violence appears to be uncommon), the sample was collected with the goal of identifying objections, not testing their relative frequencies. In particular, as described above, the sample was drawn from the top 100 comments to YouTube videos, as well as from tweets that were on topics likely to be "controversial." As intended, this sampling frame produced an abundance of objections, a necessary precondition for the codebook development.

To more formally test for these frequencies, MTurkers were recruited to code a random sample of 50 comments from one video on YouTube. As we report, we find that objections of each type are rare. In phase 3, we use a supervised classification model to assist with the identification of objections.

METHOD

Comment sampling

Phase 2 commenced roughly 8 months following the sampling of comments for the first study, so we pulled an additional video on a current news topic. Given the ease with which we could identify popular news videos on Youtube, we selected a video from one of the top 10 videos with the greatest number of comments under the "US News" category on the official CNN YouTube channel on August 16, 2022 ("CNN reporter identifies strange moment in new

Putin speech") and pulled 50 random replies to comments appended to the video. For phase 3, we similarly drew 100 random comment replies from the same video, but stratified by "stance level" as described below, with 20 comments for each of 5 levels, and then filtered these for whether they were direct replies.

Procedure

We followed the same procedure as in Study 1, recruiting and randomly assigning MTurkers to one of the seven tutorials created to introduce participants to a distinct objection strategy. Again the training consisted of informed consent, a tutorial to define an objection strategy and illustrate its use in practice (see appendix), a 6-item quality-check quiz, and for those that correctly answered at least 5 of 6 questions on the quiz, a test of 8 additional questions. Unlike in Study 1, the tests for Study 2 did not contain comments previously coded and curated, but instead included raw comments with unknown objection strategies or frequencies.

Assessments of Comments

In addition to having MTurkers code these 50 comments, we deployed two additional methods to assess them. After populating the tests in our tutorials with the random sample, our team manually coded the 50 random comments. Second, all comment replies for the video (N=5,514) were assigned a "stance" score with stance detection. Stance detection is the extraction of a subject's "stance" toward a claim made by another actor and is typically displayed as "in favor of", "against", or "neutral." This natural language processing task has been used to detect fake news, misinformation and rumors.

We used a model that incorporated tf-idf features, which is a statistical measure that evaluates how important a word is in a sentence, with Facebook's RoBERTa NLP model, which has been pre-trained on corpora containing millions of words (Prakash & Tayyar Madabushi, 2020). We ran the model on our previously coded collection of comments from the first study to determine the model's probability of predicting what we had labeled as objections. We observed that comments with a stance above 0.3 were substantially more likely to be objections.  We thus calculated stance scores for all comments in our sample pool. For the second stage of the study, we drew a random sample of 20 comments from each of 5 "bins" of stance score: 0-.1; .1-.2; .2-.3; .3-.4; .4-.5.  These comments were also coded by the team to determine "ground truth."

RESULTS

A total of 739 MTurkers were recruited and 470 (63.59%) proceeded to the test.  Quiz pass rates were consistent with those in the previous study, ranging from 41.67% to 90.22%.  Most comments were assigned to at least 4 testers. Whereas in the first study we explored the accuracy of testers to validate the training of our codebook, in this study we assess the agreement between testers.

Results for coder agreement were mostly consistent with Study 1 with one important difference. As shown in Table 4, percent agreement for each category was very high (91% or higher for each category), like it was for Study 1. However, reliability tended to be lower, with scores in the .15-.40 range for most categories.  This could be explained, statistically, by the possibility that objections in these samples were quite rare.  Krippendorff's alpha awards less "credit" for agreement on the absence of rare instances.

To investigate this possibility, we calculated the number of comments identified as cases of each category according to a supermajority rule (at least 3 out of 4 commenters agree it is an example).  As shown in Table 3, comments that use objection strategies are fairly rare: no category had more than 8 comments (16%) of the 50 total comments identified as objecting. Several categories are especially rare (4% or below), and some have no examples within the sample at all.

**Table 3.** *Results of Mturkers Coding a Non-curated Collection of Comments*

| Category | Recruitment | Final sample for test | Quiz pass | Avg. quiz score | Avg. coders[a] | Avg. agreement | Reliability[b] | Frequency |
|---|---|---|---|---|---|---|---|---|
| Label + Reject | 144 | 60 | 41.67% | 4.16 | 4.78 | 0.87 | 0.2 | 2 |
| Imploring + Logic | 139 | 59 | 42.45% | 4.26 | 4.7 | 0.88 | 0.4 | 8 |
| Imploring + Conscience | 122 | 70 | 57.38% | 4.55 | 5.7 | 0.88 | 0.46 | 5 |
| Self-preserving + Exit | 78 | 68 | 87.18% | 5.26 | 5.42 | 0.98 | 0.09 | 0 |
| Space-preserve + Removal | 87 | 70 | 80.46% | 4.99 | 5.66 | 0.92 | 0.3 | 1 |
| Smear + Undermine | 92 | 83 | 90.22% | 5.36 | 6.7 | 0.84 | 0.43 | 4 |
| Intimidate + Violence | 77 | 60 | 77.92% | 5.03 | 4.88 | 0.97 | 0.05 | 0 |

*Note.* [a]The average number of coders per comment, [b]Krippendorf's alpha was calculated.

This low frequency can account for the lower reliabilities observed in Study 2, but also presents another problem. Since coders only code 8 examples, low frequencies mean that many coders may not see any examples of the strategy that they were trained to identify. This may induce them to over-identify cases, in a sense seeing "ghost" cases that are close to what they are trained to see simply because this is what they are looking for. Consistent with this concern, we find that MTurkers were more inclined to make errors as false positives (average precision = 59%) than false negatives (average recall = 65%), though the difference is small.

To combat this "ghost" effect, we had MTurkers code a stratified sample based on stance. Our hope here was that comments with higher stance scores would be more likely to be objections, and thus by stratifying by stance, we would "boost" the number of objections to be coded, reducing the problem.  In particular, the vast majority of comments have a low oppositional stance.  In our original dataset, however, we found that objections tended to have higher stance scores (.3 or above).  If objections are mostly found among comments with high oppositional stance, oversampling from these higher strata could yield samples with a greater proportion of objections.

Results indicate that this test was not successful. We find, first, that frequencies of objection still remain low within each strata of our sample. That is, even within high oppositional stance strata objections were rare.  More importantly, we find evidence that the stratification by stance could potentially introduce selection bias. Specifically, we find that low stance strata (0-.1; .1-.2) contain the bulk of objections that use imploring with logic or conscience, while higher stance strata (.3 and higher) contain the bulk of objection strategies that smear and undermine. Thus, oversampling in higher strata would favor the identification of one kind of objection over another.  While these results are tentative, they indicate that this strategy must be adjusted before being pursued further, and we do not report any further results from it here.

DISCUSSION

User-generated comments that appear below uploaded content on social media platforms play an important role in shaping our understanding of content.  Exposure to user-generated comments appended beneath news videos can influence understanding of the news story, or impressions of related topics carried out subsequently in the discussion thread. While exposure to comments can be helpful, particularly if a user learns about a related reputable source or develops a deeper and contextualized understanding of a topic, exposure to comments can also be harmful.  When hateful comments persist, for example, or threats to stigmatized groups go unchecked, users exposed to this prejudice can become desensitized by and even accepting of it (Hsueh et al., 2015; Soral et al., 2018). Additionally, when disinformation persists or the same lie is encountered frequently within comments, users exposed to it can perceive the lie as believable (Hassan & Barber, 2021). Understanding the strategies that everyday users employ when encountering problematic and harmful content in user-generated comments informs broader understanding of the underlying mechanisms that shape user perceptions and interpretations of information encountered on social media platforms.

Findings

Our research has identified 7 distinct discursive strategies utilized in real world online discussion threads.  We find that these strategies are uncommon, creating challenges for empirical researchers. However, despite their rarity, each strategy is distinct with important differences that suggest they merit further study both in terms of how they are deployed and what their effects are.

Our findings show that users deploy a diversity of strategies, running a gamut from punishment tactics to persuasion tactics. This diversity challenges the dominant view that social media conflict is highly confrontational and full of vitriol (Polak & Trottier, 2020). We did observe a consistent category of "threatening reputation," the kind of objection most similar to popular accounts of "shaming," "flaming" or harassment.  We also observed outright rejections of content considered taboo, vicious or harmful.  However, we also observed that people often use moral principles, avoid attempts to harm another's reputation and utilize persuasion to influence behavioral changes. In these instances, users take deliberate efforts to engage constructively with another user whom they presumably do not know to bridge gaps in understanding and promote civil discourse. Users also show an awareness that social media conversations are voluntary spaces that can be managed for membership and interaction.

Sometimes, when an individual says something objectionable, they are criticized as people ("Smear + Undermine"), but other times, they are simply instructed to leave when their behaviors do not fit with the perceived norms of the space ("Space-preserve + Removal").

Limitations and Future Work

The finding of this diversity in objection strategies suggests at least two paths for further research. One path is to explore the effects of these different strategies. Our tentative results indicate that imploring with logic is more prominent than self-preserve and exit, but does it work better as a deterrent? Members of this same study team are currently exploring the efficacy of these strategies tested with an experiment. Once the efficacy of strategies is known, we anticipate the creation of education-based interventions to teach online users how to utilize these strategies in everyday online interactions. This research will add to the existing literature on digital, information and media literacy instruction, adding additional prosocial skills that individual users can take responsibility in employing.

A second path to explore is in the motivations that lead people to choose these strategies. Why do some people choose to attack the reputation of the commenter and others choose to make a more conscientious appeal? Future studies might consider surveys or interviews of individuals who have engaged in these practices to try to uncover the rationale from their point of view (Chadwick et al., 2018). Alternatively, experiments might be devised to try to nudge individuals toward choosing one strategy or the other to understand the underlying causes of these different responses.

This research also has important limitations. Most plainly, we find that objections are *rare*, creating a challenge for human coders to identify them accurately. We suspect this problem is not unique to objections. Thus, future work can contribute by identifying a more sophisticated human-in-the-loop workflow to allow human judgment to be deployed on cases where it can most accurately make determinations. Such an approach might include incentives that reflect base rates, supervised pre-filtering that goes beyond stance detection, or other innovations.

Conclusion

Research has explored the impact of intervening when harmful content infiltrates specific social settings, with studies on in-group fact-checking (Cheong, 2022; Margolin et al., 2018; Pasquetto et al., 2022), bystander programming in schools (Gaffney et al., 2019), and conspiracy theory infiltration within comments sections on state-run informational policy websites (Bradshaw et al., 2021; Sapountzis et al., 2022). In general this research has focused on the effects of such interventions, but not on nuanced distinctions across intervention strategies.

Social media companies have invested considerable time and effort in identifying, flagging and removing problematic content. While such work is important, it has not resulted in platforms that are free of hate speech or disinformation. Objectionable content remains, and individual users utilize their own discursive objection strategies to counter and silence such content. Understanding not just when people object to such comments but also how they do so in authentic ways helps advance understanding of how communities maintain and correct prosocial behaviors within comment threads.

Appendix

Objection Strategy:
**Smear + Undermine**

Definition:
Comments that attack or "smear" a user's character to undermine their authority.

Features:
Comments that do this must include an attack on at least one of the following:

- The user's intelligence (such as their education, judgment)
- The user's authenticity (accusing them of being a bot, troll)
- The user's apparent motives
- The user's principles (or apparent prejudices, selfishness)
- The user's affiliation (like political party)

Notes:

Comments that do this might or might not attack the substance or argument of the message left by the user. The most important aspect of this objection strategy is that a user's character or reputation is being attacked.

Comments that directly attack or imply fault with a person's character should be counted here, even if the "smear" doesn't feel particularly mean or slanderous to you.

Example of a tutorial in qualtrics for MTurkers